\pdfoutput=1

\documentclass[11pt]{article}

\usepackage[]{emnlp2021}

\usepackage{times}
\usepackage{inconsolata}
\usepackage{amsmath}
\usepackage{amsfonts}
\usepackage{latexsym}
\usepackage{booktabs}
\usepackage{graphicx}

\usepackage[T1]{fontenc}

\usepackage[utf8]{inputenc}

\usepackage{microtype}

\title{Assessing Data Efficiency in Task-Oriented Semantic Parsing}

\author{Shrey Desai \quad\quad Akshat Shrivastava \quad\quad Justin Rill \quad\quad Brian Moran \\ \quad\quad \textbf{Safiyyah Saleem} \quad\quad \textbf{Alexander Zotov} \quad\quad \textbf{Ahmed Aly} \\
Facebook \\
\tt{\{shreyd, akshats, jrill, bmoran,} \\\texttt{safisaleem, alexzotov, ahhegazy\}@fb.com}}

\begin{document}
\maketitle
\begin{abstract}
Data efficiency, despite being an attractive characteristic, is often challenging to measure and optimize for in task-oriented semantic parsing; unlike exact match, it can require both model- and domain-specific setups, which have, historically, varied widely across experiments. In our work, as a step towards providing a unified solution to data-efficiency-related questions, we introduce a four-stage protocol which gives an approximate measure of how much in-domain, ``target'' data a parser requires to achieve a certain quality bar. Specifically, our protocol consists of (1) sampling target subsets of different cardinalities, (2) fine-tuning parsers on each subset, (3) obtaining a smooth curve relating target subset (\%) vs. exact match (\%), and (4) referencing the curve to mine ad-hoc (target subset, exact match) points. We apply our protocol in two real-world case studies---model generalizability and intent complexity---illustrating its flexiblity and applicability to practitioners in task-oriented semantic parsing.
\end{abstract}

\begin{figure}
    \centering
    \includegraphics[scale=0.45]{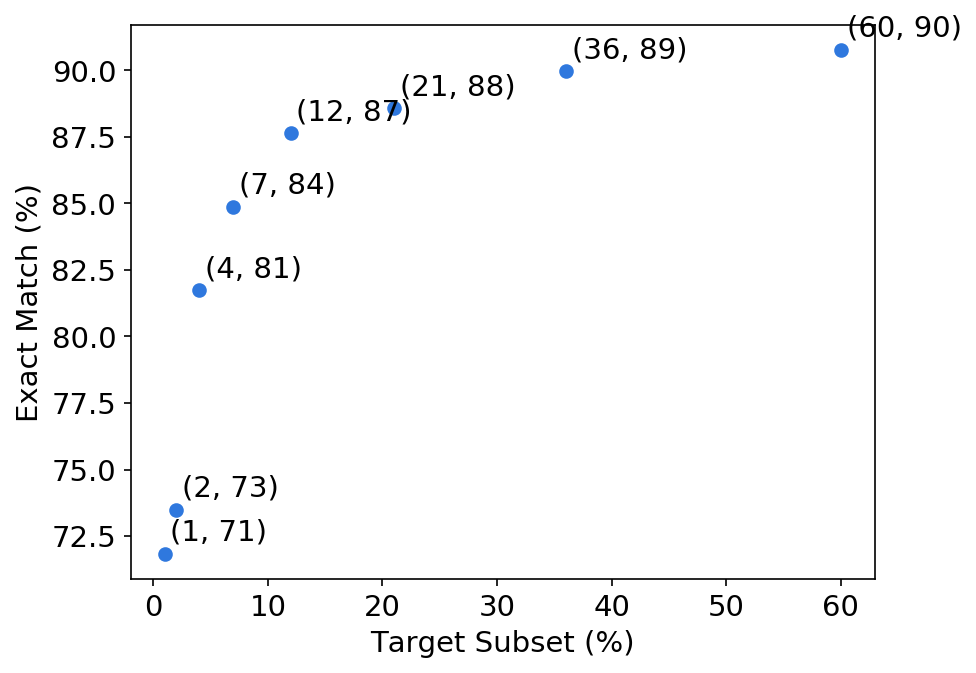}
    \includegraphics[scale=0.45]{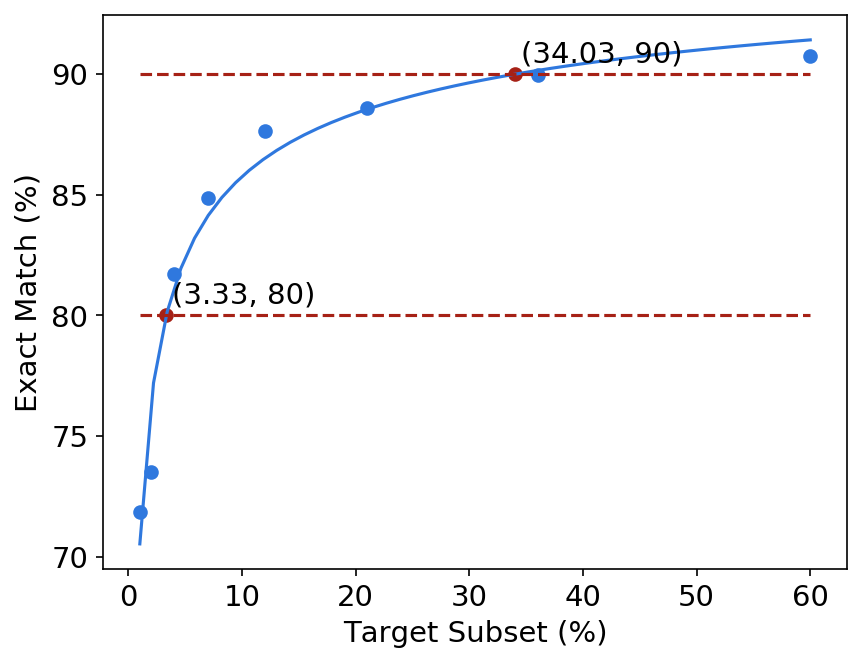}
    \caption{\textbf{Illustration of our data efficiency protocol's outputs.} The \textbf{discrete} plot (top) shows the exact match scores a task-oriented semantic parser achieves while being fine-tuned on increasingly larger subsets of (target domain) data; the logarithmic structure of this plot is typical, as model performance saturates in the presence of more in-domain data. Furthermore, the \textbf{continuous} plot (bottom) approximates the true shape of the data efficiency curve, enabling us to make ad-hoc $(x, y)$ queries; that is, how big of a $x$\% target subset does the parser require to achieve $y$\% exact match?}
    \label{fig:example-plot}
\end{figure}

\section{Introduction}

Task-oriented, conversational assistants typically first use semantic parsers to map textual utterances to structured frames executable by downstream components \cite{gupta-2018-top,einolghozati-2018-improving,pasupat-2019-span,aghajanyan-2020-decoupled,chen-2020-topv2,ghoshal-2020-loras,babu-2021-nar,desai-2021-dianosing,shrivastava-2021-span-ptr,desai-2021-intrinsic}. Because of the high costs associated with developing new conversational skills, there has been a surge of interest in improving the data efficiency of these parsers to bootstrap learning in low-resource settings \cite{chen-2020-topv2,ghoshal-2020-loras,desai-2021-intrinsic}. However, despite considerable progress in improving training, practitioners do not yet have a agreed-upon protocol to measure and optimize data efficiency in production settings \cite{chen-2020-topv2,desai-2021-intrinsic}. Partly, this is due to the lack of straightforward methodology: unlike metrics like exact match which only require executing boolean checks across system and reference frames, metrics for data efficiency require model- or domain-based setups with numerous design decisions \cite{chen-2020-topv2,desai-2021-intrinsic}.

In this work, we introduce a simple but effective protocol to assess the data efficiency of task-oriented semantic parsers in production settings. Depicted in Figure~\ref{fig:example-plot}, we design our protocol to relate target subset (\%) vs. exact match (\%); put another way, \textbf{we approximate the number of in-domain samples (from a target domain) required to achieve a quality bar}. Our protocol requires four steps: (1) Uniformly sampling subsets from a target domain, each with different cardinality (e.g., 10\%, 20\%, 30\%, etc.); (2) Fine-tuning parsers on a mix of source domain samples (out-of-domain) and subsetted, target domain samples (in-domain), then recording in-domain exact match; (3) Smoothly approximating the target subset (\%) vs. exact match (\%) curve with least squares regression; and (4) For a quality target of interest (e.g., exact match of 90\%), computing the corresponding number of samples to achieve it (e.g., target subset of 34.03\%).

We develop our protocol to be flexible in hyperparameters yet opinionated in design, seeking to meet a range of practitioners' needs along the theme of data efficiency. As such, we leverage our protocol by conducting both prescriptive and descriptive case studies on model generalization and intent complexity, respectively. Our first case study, model generalization, compares the data efficiency of multiple task-oriented semantic parsers. Our second case study, intent complexity, evaluates the correlation between data efficiency and intent complexity (i.e., how challenging the intent is to model). Across both studies, results show that our protocol can be used as-is with minimal study-specific processing, illustrating its real-world applicability to practitioners in task-oriented semantic parsing.

\section{Data Efficiency Protocol}

Our goal is to devise a model-based data efficiency protocol which can complete the following statement: \textbf{Model $m$ requires $x\%$ of samples from target domain $d$ for fine-tuning in order to achieve $y\%$ exact match.} Here, we make an important distinction between \textbf{source} and \textbf{target} domains: source domains represent high-resource settings (10K-100K samples) while target domains represent low-resource settings (1K-10K samples). While we assume access to source domains for the purposes of bootstrapping, we are primarily interested in quantitatively measuring our parsers' abilities to generalize to target domains.

\paragraph{Overview.} We begin with a high-level overview of our data efficiency protocol; Figure~\ref{fig:example-plot} shows examples of our protocol's outputs. Here, we create 8 randomly-sampled subsets consisting of $x$\% of data from a target domain, mimicking low-resource settings where in-domain is scarce. We then fine-tune our parsers on a combination of source and target samples, once for each subset, then record exact match scores on the test set of the target domain. The resulting \textbf{discrete} plot (top) resembles a logarithmic curve: at small subset sizes, exact match improves, but at large subset sizes, exact match largely saturates. Using this intuition, we fit a polynomial function to the $(x, y)$ coordinates and subsequently obtain a \textbf{continuous} plot (bottom) approximating the target subset (\%) vs. exact match (\%) relationship. 

This ultimately allows us to complete our statement; we can now query arbitrary target subset sizes required to achieve pre-defined exact match targets. For example, using our plot, we can deduce we roughly require 2.14\% of target domain data in order to achieve 80\% exact match.

\paragraph{Methodology.} In summary, our protocol consists of four steps:
\begin{enumerate}
    \item For a target domain, using a random sampling algorithm, create various training subsets, each with a different cardinality (e.g., 10\%).
    \item Fine-tune parsers on a concatenation of source domain samples (out-of-domain) and subsetted, target domain samples (in-domain). Then, report their exact match perforamnce on the test set of the target domain. (Here, source domain samples are largely used for bootstrapping and do not have overlap with target domain samples.)
    \item Obtain a smooth, continuous approximation of the target subset (\%) vs. exact match (\%) curve, as the previous step only logs discrete points.
    \item For a pre-defined exact match (\%) target (e.g., 90\%), compute the corresponding target subset (\%) required to achieve it.
\end{enumerate}

Below, we elaborate on the technical details behind each step of our protocol.

\subsection{Background}

In task-oriented semantic parsing, we are typically given a dataset $D$ comprised of the union of multiple sub-datasets $\{D^{(1)}, \cdots, D^{(d)}\}$, where each dataset $D^{(i)} = \{(x^{(j)}, y^{(j)})\}_{j=1}^n$ consists of utterance/frame pairs from domain $d$. TOPv2 \cite{chen-2020-topv2}, a popular open-source dataset, consists of the alarm, event, messaging, music, navigation, reminder, timer, and weather domains.

\subsection{Sampling Target Subsets}

Recall that we would like to assess the data efficiency of a parser in a \textbf{target} domain once it has been bootstrapped on non-target domains, or simply put, \textbf{source} domains. To achieve this, we create a ``data efficiency curve'' (as depicted in Figure~\ref{fig:example-plot}) demonstrating how well a parser extends a cross-domain setting upon being fine-tuned on increasingly larger subsets of target domain data.

\paragraph{Random Sampling Algorithm.} We create these subsets by randomly sampling utterance/frame pairs from the target domain. Specifically, our random sampling algorithm algorithm is a function $f$ which creates a subset $S^{(k_i)}_d = f(D^{(d)}, k_i)$ with size $k_i$. We parameterize $f$ as a uniform algorithm which selects $k_i\%$ of samples from $D^{(d)}$ without replacement. This algorithm is simple and well-defined: it requires minimal operations to implement and we can easily compute subset sizes as $|S^{(k_i)}_d| = \frac{1}{k_i} \times |D^{(d)}|$. See Section~\ref{sec:uniform-vs-spis} for an in-depth discussion on the advantages and disadvantages of uniform sampling.

\paragraph{Selecting Subset Sizes.} Using our random sampling algorithm $f$, we create $n$ subsets $\{S^{(k_1)}_d, \cdots, S^{(k_n)}_d\}$ for training and evaluation (discussed next). If $n$ is too small, we may not have enough points to estimate a data efficiency curve, but if $n$ is too large, our protocol will become cost-prohibitive as fine-tuning typically requires multiple GPUs. To set $n$ and $(k_1, \cdots, k_n)$ precisely, we lean on intuition from the previous section; because exact match logarithmically improves as subsets increase in cardinality \cite{chen-2020-topv2,desai-2021-intrinsic}, we can precisely capture data efficiency by selecting subset sizes along such a curve.

We develop a heuristic which empirically works well for a range of domains. We set $n = 10$ such that $k_{1} = 0$ and $k_{10} = 100$; put another way, we sample 10 subsets where the 1st and 10th subsets have 0\% and 100\% of target domain data, representing the lower and upper bounds, respectively. The remaining 8 subsets are spaced out along a logarithmic curve to mimic the typical characteristics of cross-domain generalization.

Formalizing this process, we seek to build a function $g(x)$ with domain $x \in [1, 10]$ and range $g \in [0, 100]$ such that $g$ is spaced out logarithmically and $k_i = \lceil g(i) \rceil$ for ease of use. We can use the generic function $g(x) = a^{x-b} + c$ as a template with the additional constraints that $g(1) = 0$ (0\% subset) and $g(10) = 100$ (100\% subset). With some algebra, we can solve for the unknown variables, building the following function:

\begin{equation}
    g(x) = (\sqrt[9]{101})^{x-1} - 1
\end{equation}

We can now determine $k_{1:10}$ easily by evaluating $g$ over its domain; Table~\ref{tab:exponential-function} shows the inputs and outputs of this function for reference. Therefore, using uniform sampling, we create 10 subsets with 0\%, 1\%, 2\%, 4\%, 7\%, 12\%, 21\%, 36\%, 60\%, and 100\% of target data.

\begin{table}[]
\centering
\begin{tabular}{lrr}
\toprule
$x$ & $g(x)=a^{x-b}-1$ & $g(x)=\lceil a^{x-b}-1 \rceil$ \\
\midrule
1 & 0.00 & 0 \\
2 & 0.67 & 1 \\
3 & 1.79 & 2 \\
4 & 3.66 & 4 \\
5 & 6.78 & 7 \\
6 & 11.99 & 12 \\
7 & 20.69 & 21 \\
8 & 35.22 & 36 \\
9 & 59.48 & 60 \\
10 & 100.00 & 100 \\
\bottomrule
\end{tabular}
\caption{We create an exponential function $g(x) = (\sqrt[9]{101})^{x-1} - 1$ (where $a = \sqrt[9]{101}$ and $b = 1$) to determine target subset sizes for uniform sampling. Note that we use the ceiling function to discretize the output space of our function.}
\label{tab:exponential-function}
\end{table}

\subsection{Training and Evaluation}

Having covered how subsets of target domain data are created, we now describe how to populate the discrete points of a data efficiency curve. Because we plot target subset (\%) vs. exact match (\%), as shown in Figure~\ref{fig:example-plot}, we \textbf{independently} fine-tune the same parser $n$ times on train/eval data---a mix of source and (subsetted) target data---then subsequently report its exact match on the target domain's corresponding test set.

Using the notation from the previous section, for each target subset $S^{(i)}_d$, we create training and evaluation datasets $\{D^{(d')} | d' \ne d\} + \{S^{(i)}_d\}$ for fine-tuning. Here, note that although we homogenize source and target data, we are only interested in target domain test performance, as the source domain largely acts as a bootstrap for cross-domain generalizability. 

Once we perform fine-tuning $n$ times, once for each target subset, we create a \textbf{discrete} plot with two axes where the $x$-axis represents target subset (\%) and the $y$-axis represents exact match (\%). For example, in Figure~\ref{fig:example-plot}, we see that the parser achieves 70\% EM with 1\% of target data but quickly improves to 82\% EM with 3\% of target data.

\subsection{Continuous Approximation}

The discrete plot gives a coarse picture of data efficiency, but the target subset  (\%) vs. exact match (\%) relationship is only defined for the subset sizes we previously selected. To define this relationship for \textit{all} possible subset sizes in a cost-effective fashion, one possible solution is creating a \textbf{continuous} plot; because we have numerous, logarithmically-spaced points, we can fit a continuous function on these points which effectively ``fills in the gaps''. Note that although this solution is approximate rather than exact, we empirically find our method accounts for variance between fine-tuning runs and are reasonably accurate due to our heuristic selection of initial subset sizes.

For curve fitting, we select the general-purpose polynomial function $h(x) = \frac{a}{x^b} + c$ with parameters $\theta = [a, b, c]$ for its simplicity and flexibility. We use \texttt{scipy.optimize.curve\_fit}, which learns $\theta$ by minimizing a least squares objective. For reference, Figure~\ref{fig:example-plot} shows a comparison of a discrete and continuous plot.

\subsection{Samples vs. EM}

Once we have learned a continuous function $h(x)$ with parameters $\theta$ over our discrete points, we can now achieve our original goal; that is, evaluating the number of samples required to achieve a certain exact match. We create an inverse continuous function $h^{-1}(y) = (\frac{y-c}{a})^\frac{-1}{b}$ and pass in pre-defined exact match targets of interest (e.g., $y=90$ for 90\% exact match). The output of $h^{-1}(y)$ can be interpreted as the $x$ such that $h(x) = y$, or in other words, the target subset (\%) which results in an exact match (\%). Furthermore, optionally, we can recover the exact number of samples here ($\frac{1}{x} \times |D^{(d)}|$) as target subsets are uniformly sampled.

Figure~\ref{fig:example-plot} illustrates the procedure described above. By fitting our polynomial function to the discrete points, we end up with the canonical function $h(x) = \frac{-27.26}{x^{0.35}} + 97.79$ and inverse function $h^{-1}(y) = (\frac{y-97.79}{-27.26})^\frac{-1}{0.35}$. From here, we can plug in pre-defined $y$ values of interest; for example, $h^{-1}(80) \approx 3.33$ and $h^{-1}(90) \approx 34.03$ as shown by the dotted red lines in the bottom plot.

\section{Uniform vs. SPIS}
\label{sec:uniform-vs-spis}

One important design decision we make in our protocol is using a uniform algorithm---as opposed to samples per intent slot (SPIS) \cite{chen-2020-topv2}---to sample target subsets. Similar to the uniform algorithm, the SPIS algorithm is a function parameterized by a sizing parameter $k$. However, instead of selecting a percentage of samples from $D^{(d)}$, SPIS builds up a subset $S_d^{(k_i)}$ which consists of at least $k_i$ occurrences of \textit{each} ontology label (i.e., intent or slot).

We discuss a couple of disadvantages of the SPIS algorithm. First, because SPIS-based subset sizes are governed by label occurrences, their sizes are dynamic as opposed to static. It is not immediately clear how large a SPIS-based subset is unless it is explicitly computed; this is not the case with uniform-based subsets, as their size only depends on the total number of samples per domain. Second, also due to the label occurrences constraint, it is often challenging to select subset sizes $(k_1, \cdots, k_n)$ for fine-tuning. Some domains are very small (10-100 samples) while other domains are very large (1K-10K samples), so we would need to choose subset sizes depending on a domain's characteristics; for example, 1000 SPIS would be too large of a subset size to use in a small domain.

However, unlike SPIS, uniform sampling does not guarantee coverage over the entire ontology space, as it only samples utterance/frame pairs according to empirical frequency. This is an important consideration for closed-set domain adaptation where the output space must stay static \cite{chen-2020-topv2}. Nonetheless, for our purposes, the characteristics of uniform sampling are a ``feature'' rather than a ``bug'', as it creates subsets which more naturally reflect the underlying data distribution.

\section{Experimental Setup}

For the remainder of this paper, we shift towards leveraging our data efficiency protocol in a series of experiments, each investigating prescriptive or descriptive hypotheses practitioners may pose when building task-oriented assistants.

We experiment with a range of task-oriented semantic parsing models when conducting case studies. Each model is a seq2seq transformer with a pointer-generator-based decoder and relies on either autoregressive (AR) or non-autoregressive (NAR) generation:\footnote{Refer to \citet{shrivastava-2021-span-ptr} for training details and model hyperparameters.}
  
\paragraph{BART AR.} BART is a seq2seq transformer combining a transformer encoder and autoregressive transformer decoder, and is pre-trained with a denoising autoencoder objective on monolingual corpora \cite{lewis-2020-bart}. For task-oriented semantic parsing, \citet{aghajanyan-2020-decoupled} shows BART achieves state-of-the-art EM on multiple datasets.


\paragraph{RoBERTa NAR.} Unlike RoBERTa AR, RoBERTa NAR assumes strong independence assumptions during decoding, using the mask-predict algorithm to enable non-autoregressive generation \cite{ghazvininejad-2019-maskpredict}. We use the framework outlined in \cite{babu-2021-nar}, creating a seq2seq transformer with a RoBERTa encoder, MLP length module, and a non-autoregressive, randomly-initialized transformer decoder (1L, 768H, 16A).

\paragraph{RoBERTa Span Pointer.} Compared to RoBERTa NAR, RoBERTa Span Pointer \cite{shrivastava-2021-span-ptr} optimizes the frame representation and the model architecture to be span-based, but also relies on non-autoregressive decoding. Specifically, utterance spans are represented as index-based endpoints in frame slots (e.g., \texttt{[i, j]}), and the transformer decoder is modified accordingly to place a distribution on indices rather than words during generation.

\begin{figure*}[t]
    \centering
    \includegraphics[scale=0.5]{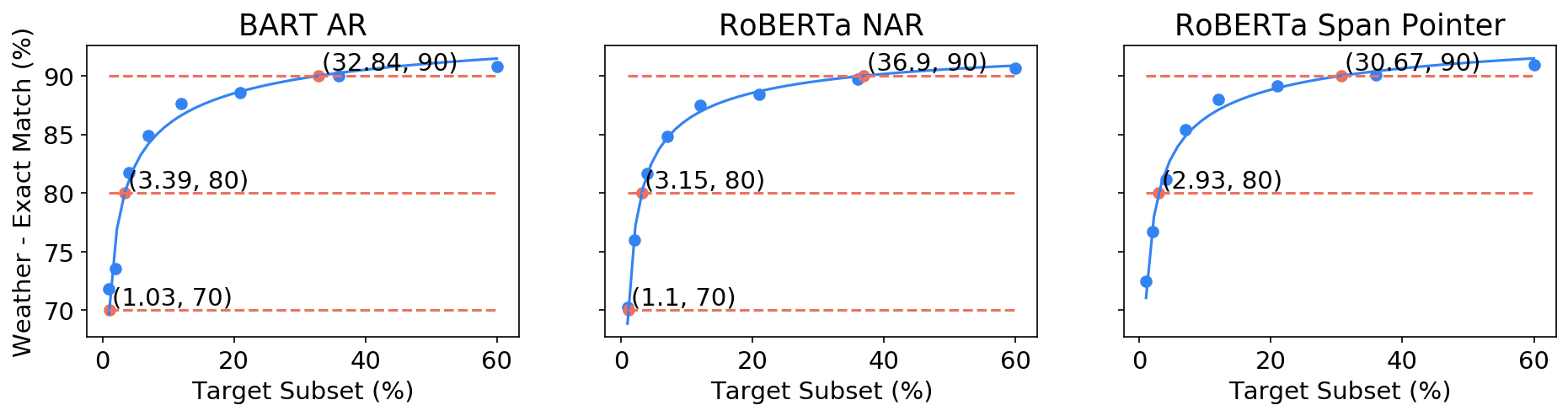}
    \includegraphics[scale=0.5]{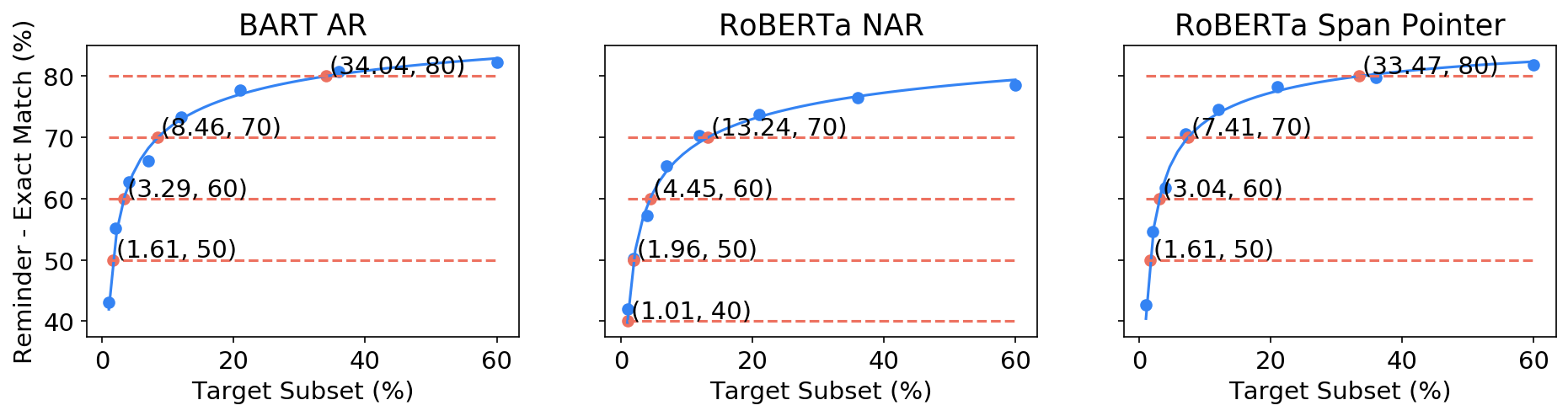}
    \caption{Data efficiency of various production-ready, transformer-based seq2seq parsers on the weather (top) and reminder (bottom) domains. The RoBERTa Span Pointer model is the most generalizable, achieving 90\% exact match with 30.67\% target data on weather and 80\% exact match with 33.47\% target data on reminder.}
    \label{fig:de-plot}
\end{figure*}

\begin{figure}
    \centering
    \includegraphics[scale=0.43]{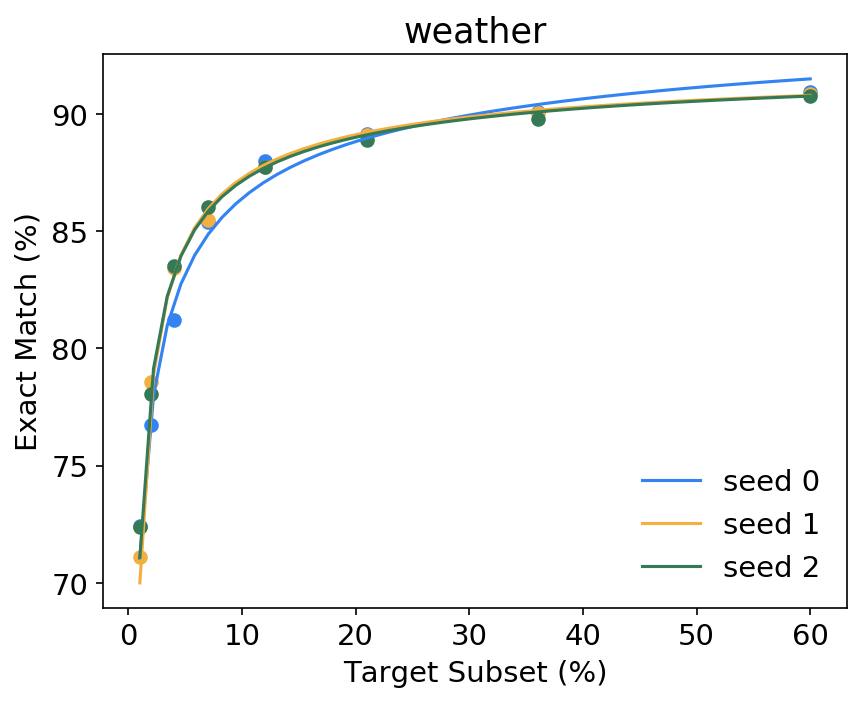}
    \includegraphics[scale=0.43]{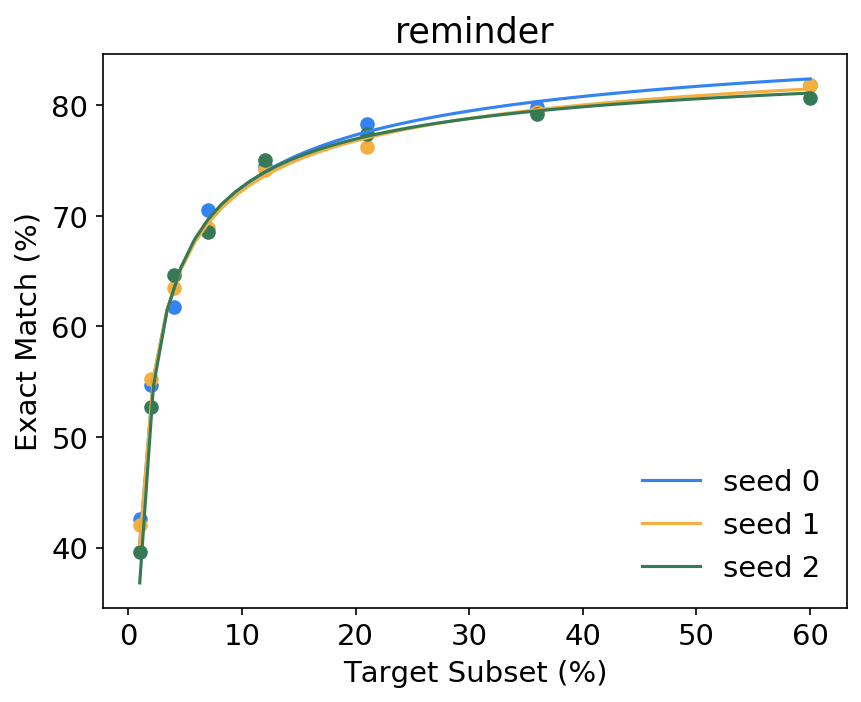}
    \caption{Demonstrating the variance of our model generalizability results by fine-tuning RoBERTa Span Pointer parsers with \{0, 1, 2\} random seeds for the weather (top) and reminder (bottom) domains. Despite fluctuations in discrete values, the continuous plots are tightly bounded together, indicating the robustness of our protocol.}
    \label{fig:variance-plot}
\end{figure}

\section{Case Study \#1: Model Generalizability}

Our first case study concerns model generalization: \textbf{How data efficient are production-ready, task-oriented semantic parsing models?} Because crowdsourcing annotations for new domains is time-consuming and cost-prohibitive, practitioners may instead opt for using data efficient parsers which perform well given a limited amount of in-domain data \cite{chen-2020-topv2,ghoshal-2020-loras,desai-2021-intrinsic}. However, given the recent explosion in the number of transformer-based semantic parsers, it can be daunting to select the one which is most generalizable. Our protocol can naturally provide an answer: by bootstrapping a parser on high-resource, source data, then exposing it to increasingly larger subsets of low-resource, target data, we can evaluate generalizability in a data-driven fashion.

Following prior work \cite{chen-2020-topv2,ghoshal-2020-loras}, we use the weather and reminder domains of TOPv2 as (independent) target domains; these domains are highly distinct, as they vary across multiple axes (e.g., utterance length, ontology size, frame nesting) \cite{desai-2021-intrinsic}.

\begin{table}[]
\begin{tabular}{ll}
\toprule
\multicolumn{2}{l}{\textbf{Domain: Music}} \\
\midrule
\texttt{IN:ADD\_TO\_PLAYLIST\_MUSIC} & open \\
\texttt{IN:CREATE\_PLAYLIST\_MUSIC} & open \\
\texttt{IN:DISLIKE\_MUSIC} & closed \\
\texttt{IN:LIKE\_MUSIC} & closed \\
\texttt{IN:LOOP\_MUSIC} & closed \\
\texttt{IN:PAUSE\_MUSIC} & closed \\
\texttt{IN:PLAY\_MUSIC} & semi \\
\texttt{IN:PREVIOUS\_TRACK\_MUSIC} & none \\
\texttt{IN:REMOVE\_FROM\_PLAYLIST\_MUSIC} & semi \\
\texttt{IN:REPLAY\_MUSIC} & closed \\
\texttt{IN:SET\_DEFAULT\_PROVIDER\_MUSIC} & closed \\
\texttt{IN:SKIP\_TRACK\_MUSIC} & closed \\
\texttt{IN:START\_SHUFFLE\_MUSIC} & closed \\
\texttt{IN:STOP\_MUSIC} & closed \\
\midrule
\multicolumn{2}{l}{\textbf{Domain: Messaging}} \\
\midrule
\texttt{IN:CANCEL\_MESSAGE} & none \\
\texttt{IN:GET\_MESSAGE} & semi \\
\texttt{IN:IGNORE\_MESSAGE} & none \\
\texttt{IN:REACT\_MESSAGE} & closed \\
\texttt{IN:SEND\_MESSAGE} & open \\
\midrule
\multicolumn{2}{l}{\textbf{Domain: Reminder}} \\
\midrule
\texttt{IN:CREATE\_REMINDER} & open \\
\texttt{IN:DELETE\_REMINDER} & open \\
\texttt{IN:GET\_RECURRING\_DATE\_TIME} & semi \\
\texttt{IN:GET\_REMINDER} & open \\
\texttt{IN:GET\_TODO} & open \\
\texttt{IN:SEND\_MESSAGE} & open \\
\texttt{IN:UPDATE\_REMINDER} & open \\
\texttt{IN:UPDATE\_REMINDER\_DATE\_TIME} & open \\
\midrule
\multicolumn{2}{l}{\textbf{Domain: Timer}} \\
\midrule
\texttt{IN:ADD\_TIME\_TIMER} & closed \\
\texttt{IN:CREATE\_TIMER} & closed \\
\texttt{IN:DELETE\_TIMER} & none \\
\texttt{IN:GET\_TIME} & semi \\
\texttt{IN:GET\_TIMER} & none \\
\texttt{IN:PAUSE\_TIMER} & none \\
\texttt{IN:RESTART\_TIMER} & none \\
\texttt{IN:RESUME\_TIMER} & none \\
\texttt{IN:SUBTRACT\_TIME\_TIMER} & closed \\
\texttt{IN:UPDATE\_TIMER} & closed \\
\midrule
\multicolumn{2}{l}{\textbf{Domain: Weather}} \\
\midrule
\texttt{IN:GET\_SUNRISE} & semi \\
\texttt{IN:GET\_SUNSET} & semi \\
\texttt{IN:GET\_WEATHER} & semi \\
\bottomrule
\end{tabular}
\caption{Intent complexity annotations for the music, messaging, reminder, timer, and weather domains following the complexity classes none, closed, (semi)-open, and open.}
\label{tab:annotations}
\end{table}

\subsection{Results}

Figure~\ref{fig:de-plot} shows data efficiency plots for models fine-tuned on the weather and reminder domains. We highlight a couple of key trends:

\paragraph{Autoregressive parsing is more data efficient than canonical non-autoregressive parsing.} We see that BART AR is more data efficient than RoBERTa NAR; on weather, to achieve 90\% EM, BART AR requires 32.85\% of target data while RoBERTa NAR requires 36.90\% of target data, and on reminder, to achieve 70\%, BART AR requires 8.46\% of target data while RoBERTa NAR requires 13.24\% of target data. Inspecting these results closer, we find the length module in RoBERTa NAR is a major bottleneck for generalization; because of strong conditional independence assumptions during non-autoregressive decoding, this parser must first predict the length of the frame to later infill, which can be challenging in a few-shot setting. In contrast, autoregressive parsing does not require an intermediate step and therefore is much simpler to extend cross-domain.

\paragraph{However, span-based, non-autoregressive parsing is highly data efficient.} RoBERTa Span Pointer, despite being a non-autoregressive parser, achieves the best data efficiency results compared to both BART AR and RoBERTa NAR. Recall that this model carries a different inductive bias as it reformulates parsing to be span-based: utterance spans in leaf arguments are represented as index-based endpoints rather than string-based text. This model's length module no longer has to guess the length of leaf arguments beforehand and can instead focus on the (predictable) syntactic components (e.g., the number of ontology tokens); as a result, both RoBERTa Span Pointer's length prediction accuracy and final exact match is substantially greater than RoBERTa NAR's. \textbf{Given RoBERTa Span Pointer's strong data efficiency results, we recommend practitioners use this model in production settings.}

\paragraph{Data efficiency results do not change much between fine-tuning runs.} Because our protocol is approximate rather than exact, one question we investigate is how stable our protocol's results are between fine-tuning runs. Here, we are interested in two types of variation: the change in discrete points when when different random seeds are used and, subsequently, the change in continuous curves fitted on each set of discrete points. We fine-tune the RoBERTa Span Pointer model with random seeds \{0, 1, 2\} and create data efficiency plots using our protocol; Table~\ref{fig:variance-plot} shows these results. For both the weather and reminder domains, we see that our data efficiency plots are quite similar a cross fine-tuning runs. Though the discrete points typically fluctuate by $\pm$1 EM, as is expected due to different random initializations, the continuous curves are largely similar. These curves can be tighter given more discrete points, especially at larger target subset sizes, but it inevitably comes at the cost of using more compute.

\begin{figure*}
    \centering
    \includegraphics[scale=0.5]{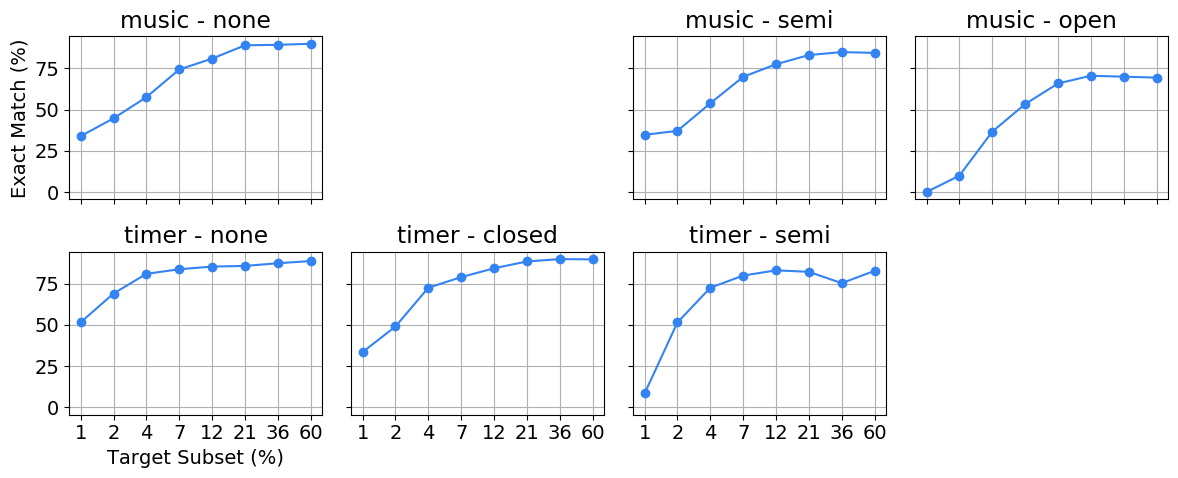}
    \caption{Comparing data efficiency with intent complexity for the music (top) and timer (bottom) domains. For each domain, we show a discrete plot for the four complexity classes: none, closed, (semi)-open, and open; empty plots imply the domain does not have an intent annotated with the corresponding complexity class.}
    \label{fig:complexity-plot}
\end{figure*}

\section{Case Study \#2: Intent Complexity}

Our second case study concerns intent complexity: \textbf{Given a rough estimate of how ``complex'' an intent is (from a modeling standpoint), do we see a correlation between complexity and data efficiency?} When developing new domains for task-oriented assistants with an intent-driven methodology\footnote{Here, the practitioner creates an intent (e.g., \texttt{IN:SEND\_MESSAGE}), enumerates the possible slots (e.g., \texttt{SL:CONTACT\_NAME}, \texttt{SL:DATE\_TIME}, etc.), then obtains samples for modeling.}, practitioners have a range of tools at their disposal to obtain data---from crowdsourcing unique samples or paraphrasing existing samples---but it is often unclear how many samples a parser requires to achieve high quality. Here, the complexity of an intent, defined below precisely, can serve as a heuristic, as more challenging intents will subsequently require more data. However, this rests on the assumption that there is a correlation between complexity and data efficiency. Our protocol can provide a solution: once we create data efficiency plots per-intent, we can create an ``average'' plot representing the intents from each complexity class, then visually inspect the shape of these plots; if the assumption holds, we should see these plots shift towards lower exact match (\%) scores as complexity increases.

\subsection{Complexity}

\paragraph{Definition.} We allude to the notion of ``complexity'' above, but we have not yet precisely defined this term. \textbf{In our study, complexity is a rough notion of how difficult an ontology label is to model as judged by exact match.} This notion is model-agnostic: we can expect parsers, as a whole, to struggle with certain types of labels (e.g., open-text slots), even if some parsers are comparatively more accurate than others. Specifically, we define four complexity classes: (1) \textbf{none:} variable, consisting of no values, but because out-of-domain intents fall under this category, they can be more challenging to model; (2) \textbf{closed:} easy, consisting of roughly ~10 values; (3) \textbf{semi-open:} medium, consisting of named entities, date-times, or closed class with roughly ~100 values; and (4) \textbf{open:} hard, consisting of long free text.

Because our goal here is to correlate intent complexity with data efficiency, we primarily focus on annotating intents with closed, semi-open, and open classes. \textbf{To do so, we make the assumption that the complexity of an intent is derived from the maximum complexity of its slots.} This is not a particularly strong assumption, as intents and slots are strongly intertwined during modeling. For example, the intent \texttt{IN:SEND\_MESSAGE} is ``open'' since its slot \texttt{SL:CONTENT\_EXACT} is also ``open''; this particular slot maps to constituents with wide syntatic and semantic variation, therefore the overarching intent is also challenging to model.

\paragraph{Annotation.} Using our definition of complexity, we select 5 TOPv2 domains for annotation: music, messaging, reminder, timer, and weather. We have 2 in-house linguists annotate each intent from each domain with its respectively complexity class; however, intents with less than 10 occurrences are excluded from our analysis. Our linguists agree on an annotation guideline beforehand and also jointly resolve tricky cases, but for the purposes of quality estimation, both linguists blindly annotate intents in the weather domain; annotator agreement is perfect as measured by Krippendorff's alpha \cite{krippendorff-2004-alpha}. Table~\ref{tab:annotations} shows these results.

\subsection{Results}

We primarily experiment with the music and timer domains given their intents span a wide range of complexity classes. To correlate data efficiency with intent complexity, we use our protocol on the RoBERTa Span Pointer parser to obtain discrete plots for each domain. Because we are interested in intent complexity, we additionally break down the target subset (\%) vs. exact match (\%) results by intent. Then, we group the intents in each domain by their complexity class and average their results to obtain a discrete plot for each complexity class. Figure~\ref{fig:complexity-plot} shows these results. \textbf{For both the music and timer domains, we see a rough correspondence between data efficiency and intent complexity}; specifically, as the complexity increases, the discrete plots' heads tend to flatten, indicating the parser performs worse with less data. Because TOPv2 has a limited number of intents, it is challenging to create guidelines using these results, but our breakdown helps give a sense of how much in-domain data, on average, is required to achieve strong performance.

\section{Conclusion}

We introduce a data efficiency protocol for task-oriented semantic parsing capable of producing discrete and continuous plots with target subset (\%) vs. exact match (\%); this gives us both exact and approximate results, respectively, illustrating the amount of in-domain data a parser requires to achieve a quality bar. To demonstrate its real-world applicability to practitioners, we leverage our protocol in two case studies: the first study compares the data efficiency of production-ready parsers while the second study correlates data efficiency with intent complexity for the purposes of developing heuristics for data collection.

\bibliography{anthology,custom}
\bibliographystyle{acl_natbib}

\newpage
\appendix

\end{document}